\documentclass[runningheads]{llncs}
\usepackage[T1]{fontenc}
\usepackage{graphicx}
\usepackage{bm}
\usepackage{amsfonts}
\usepackage{amsmath}
\usepackage{algorithm,algpseudocode}
\usepackage{svg}
\usepackage{adjustbox}
\usepackage{hyperref}

\begin{document}
\title{Partition-Guided Distance Saliency: Bridging Decision and Objective Spaces in Many-Objective Optimization}
\titlerunning{PGDS: Bridging Decision and Objective Spaces in MaO}
%

\author{Cláudio Lúcio do Val Lopes\inst{1,2}\orcidID{0000-0003-1655-2283} \and
Flávio Vinícius Cruzeiro Martins\inst{2}\orcidID{0000-0002-6666-653X} \and
Elizabeth Fialho Wanner\inst{2,3}\orcidID{0000-0001-6450-3043}}
\authorrunning{Cláudio L. Lopes et al.}
%
\institute{A3Data, BH MG Brazil \\
\email{claudio.lucio@a3data.com.br}\\
\url{www.a3data.com.br} \and
CEFET-MG - Centro Federal de Educação Tecnológica de Minas Gerais, BH MG Brazil \\
\email{flaviocruzeiro@cefetmg.br} \and
Aston University, Birmingham, West Midlands, U.K. \\
\email{e.wanner1@aston.ac.uk}
}

\maketitle              
\begin{abstract}

Explainability in Many-Objective Optimization (MaO) is currently hindered by the escalating complexity of the Pareto front, which renders the relationship between high-dimensional decision variables and objective outcomes increasingly opaque. As the number of objectives exceeds the limits of traditional visualization, decision-makers encounter a ``cognitive drought'' in identifying relevant trade-offs or specifying target regions without \textit{a priori} knowledge. To bridge this interpretability gap, we introduce the {Partition-Guided Distance Saliency (PGDS)} framework, a novel XAI approach designed for continuous optimization landscapes. 
Our framework automates the explanation process through a three-stage pipeline that prioritizes geometric intuition over abstract rules. First, we employ a surrogate model that learns how geometric distances in the decision space map to proximity in the objective space. Second, to address the difficulty of manual target selection in high dimensions, the framework automatically partitions the objective landscape into distinct regions and identifies local ``Dominating Points'' to serve as automated targets for improvement. Third, we quantify how sensitive a solution's position is to each decision variable by measuring the distance shifts induced by perturbations to each variable. This allows PGDS to categorize features as either ``Drivers'' which facilitate convergence toward preferred regions, or ``Blockers'' which represent geometric constraints hindering further progress. Validation on 10-objective benchmarks and a physics-informed engineering problem (Welded Beam) demonstrates that PGDS provides differentiated, actionable insights that traditional visualization and rule-based XAI methods fail to provide.

\end{abstract}

\section{Introduction}\label{sec:introduction}

In the realm of Multi-Objective Optimization (MOO), the simultaneous optimization of conflicting criteria yields not a single optimal solution but a set of trade-off solutions, known as the Pareto-optimal set. 
As the number of objectives increases beyond three, a scenario commonly referred to as Many-Objective Optimization (MaO), the complexity of the problem space grows exponentially. Evolutionary Algorithms (EAs), such as the Non-dominated Sorting Genetic Algorithm (NSGA-III) \cite{deb2014nsga} and the Multi-Objective Evolutionary Algorithm based on Decomposition (MOEA/D) \cite{zhang2007moead}, have proven highly effective in approximating these Pareto fronts \footnote{Pareto front represents the mapping of the Pareto-optimal set onto the objective space}. However, finding an approximation of the Pareto front is merely the first step in a broader decision-making process. The ultimate goal is to assist a Decision Maker (DM) in selecting a single, most preferred solution from this potentially vast set \cite{miettinen1999nonlinear}.

In low-dimensional objective spaces (two or three objectives), visualization techniques such as scatter plots or parallel coordinates can sufficiently aid the DM in understanding trade-offs. However, in many-objective scenarios, these traditional visualization methods break down, rendering the relationship between the \textit{decision space}, the variables controlled by the DM, and the \textit{objective space} increasingly opaque \cite{tusar2015visualization}. A significant gap exists in providing the DM with intuitive, explainable insights into \textit{why} a specific solution resides in a particular region of the objective space and \textit{how} perturbations in the decision variables influence its position relative to desired targets.

Current research in Explainable AI (XAI) for optimization often relies on rule-based machine learning models to classify solutions or extract logical predicates \cite{misitano2022,misitano2024exploring}. While effective for discrete categorization, these methods frequently require discretizing continuous landscapes, thereby losing the geometrical fidelity necessary for precise engineering design or complex resource allocation, for example. Furthermore, they often place a heavy cognitive burden on the DM to define specific ``interesting'' regions a priori, a task that is non-trivial in high-dimensional spaces.

To bridge this gap, this paper addresses the following research questions:
\begin{description}
    \item [RQ1:] How can we bridge the interpretability gap between decision variables and objective outcomes in continuous, many-objective landscapes without losing geometric fidelity?
    \item [RQ2:] How can we automate the discovery of ``regions of interest'' in high-dimensional objective spaces to guide the DM, rather than relying solely on manual target specification?
    \item [RQ3:] Can a distance-based surrogate model effectively quantify the sensitivity of objective space positioning to decision variable perturbations?
\end{description}

To answer these questions, we propose a novel framework: Partition-Guided Distance Saliency (PGDS). This approach decouples the concept of explainability from rigid rule-based classifiers \cite{misitano2024exploring}, adopting instead a continuous, geometric perspective that respects the true geometry of the many-objective landscape. Our method employs the Minimal Learning Machine (MLM) \cite{deSouza2013mlm} to construct a regression model based entirely on distance matrices. This offers a computationally efficient mapping, scaling with $\Theta(K^2N)$, where $N$ represents the total number of solutions in the archive and $K$ denotes the subset of reference points used to anchor the geometric structure. By integrating this geometric mapping with automated region discovery, we enable the mapping between the geometry of the input (decision variables) and output (objective functions) spaces.

Central to our proposal is the adaptation of the \textit{Distance Explainer} methodology, originally developed to interpret deep learning embeddings \cite{meijer2025explainable}. We extend this concept to tabular decision variables in optimization, allowing us to generate saliency maps that quantify how specific decision variables influence a solution's geometric proximity to a target solution in the objective space. To address the difficulty of selecting targets in many-objective spaces, we introduce an automated partitioning strategy based on K-Dimensional Trees (KD-Trees) \cite{bentley1975multidimensional}. This strategy partitions the objective space into hyper-rectangular regions and identifies ``dominant points'' within each block as automatic targets for explanation.

The remainder of this paper is structured as follows: Section \ref{sec:related_work} reviews related work on interactive MOO/MaO and the current explainability approaches. Section \ref{sec:methodology} details our theoretical framework, including the MLM formulation, KDTree partitioning, and the adaptation of the Distance Explainer. Section \ref{sec:experiments} presents experimental validation on synthetic benchmarks (DTLZ2, WFG3) and a constrained engineering problem. Finally, Section \ref{sec:conclusion} concludes the paper and outlines future research directions.

\section{Related work}\label{sec:related_work}

This section reviews the existing literature on multi- and many-objective optimization, the current state of Explainable AI within the optimization domain, and the emerging class of distance-based learning methods. We frame this discussion around the three core research questions elicited in Section \ref{sec:introduction} reproduced here for the reader's convenience: (RQ1) bridging the interpretability gap without losing geometric fidelity, (RQ2) automating the discovery of regions of interest, and (RQ3) using distance-based surrogates for sensitivity analysis.

\subsection{Complexity and Visualization in Many-Objective Optimization}

Multi- and many-objective optimization involves the simultaneous optimization of a vector of typically conflicting objectives $F(x) = (f_1(\bm{x}), \dots, f_M(\bm{x}))$. Fundamental to solving these problems is the concept of Pareto dominance: a solution $\bm{x_u}$ is said to dominate another solution $\bm{x_v}$ (denoted as $\bm{x_u} \prec \bm{x_v}$) if $\bm{x_u}$ is no worse than $\bm{x_v}$ in all objectives and strictly better in at least one \cite{miettinen1999nonlinear}. The ultimate goal of an optimizer is to approximate the Pareto Front (PF), the set of all non-dominated solutions in the objective space representing the optimal trade-offs.

For problems with two or three objectives, algorithms such as the Non-dominated Sorting Genetic Algorithm II (NSGA-II) \cite{deb2002nsga2} have established themselves as the standard, effectively maintaining diversity while converging toward the PF. However, real-world problems often scale to four or more objectives, a domain known as many-objective optimization. In such high-dimensional spaces, the selection pressure of standard dominance-based methods vanishes as nearly all solutions become non-dominated. Consequently, specialized algorithms like NSGA-III \cite{deb2014nsga} and MOEA/D \cite{zhang2007moead} were developed to handle these landscapes using reference-point adaptation and decomposition strategies, respectively.

However, generating a well-converged approximation of the PF addresses only the computational aspect; the decision-making bottleneck remains. As noted by Tušar and Filipič \cite{tusar2015visualization}, traditional visualization techniques lose effectiveness beyond three dimensions. To cope with this, practitioners often rely on dimension reduction techniques adapted for high-dimensional Pareto fronts. These include mapping solutions using Principal Component Analysis (PCA) to identify principal conflict directions \cite{zhen2017objective}, applying t-Distributed Stochastic Neighbor Embedding (t-SNE) to cluster trade-off regions \cite{zou2019visual}, or using Self-Organizing Maps (SOM) to preserve the topological structure of the trade-off surface in a lower-dimensional grid \cite{NAGAR2023101202}. While useful for clustering, these methods inevitably introduce distortions when projecting high-dimensional data into 2D \cite{fieldsend2016visualising}. Such projections often fail to preserve global distance relationships or local topology, rendering the relationship between the decision variables (design space) and the resulting objective values (performance space) opaque. This disconnection highlights the need for explainability methods that operate directly on the high-dimensional geometry rather than on distorted projections.

\subsection{Explainable Optimization and the Target Selection Bottleneck}

To bridge the gap between opaque algorithmic results and human understanding, the field of Explainable AI (XAI) has recently expanded into Evolutionary Computation. While standard XAI methods such as SHAP \cite{lundberg2017unified} and LIME \cite{ribeiro2016should} successfully interpret prediction models by quantifying feature contributions, applying these concepts to optimization introduces distinct challenges. Unlike classification, where the output is a discrete class, optimization involves selecting a decision vector from a continuous, potentially infinite feasible space. Consequently, recent literature distinguishes between \textit{explaining the optimization process}, diagnosing algorithmic dynamics and operator contributions \cite{abed2026evomapx}, and \textit{explaining the solution}, providing rationale for why a specific trade-off is optimal or preferred \cite{zhou2024evolutionary}.

Our work aligns with the latter category, focusing on the post-hoc analysis of the final Pareto front. Several data-driven frameworks have emerged to address this.  In the context of Interactive Multi-Objective Optimization, the R-XIMO framework \cite{misitano2022} adapts SHAP values to quantify how a Decision Maker's aspiration levels (reference points) positively or negatively impact the achieved objective values. Similarly, the XLEMOO framework \cite{misitano2024exploring} employs a learning mode to extract logical rules (e.g., IF-THEN predicates) that characterize high-quality regions of the search space.

However, significant barriers remain when applying these methods to many-objective problems (addressing RQ1). Rule-based classifiers often require discretizing continuous variables, potentially sacrificing the fine-grained geometric fidelity required for precision engineering. Furthermore, interactive frameworks rely heavily on the DM to articulate initial preferences or select reference points \cite{miettinen1999nonlinear}. In high-dimensional MaO scenarios ($M >> 3$), DMs often encounter a ``cognitive drought'' unable to intuitively visualize the landscape or specify where to look. This creates a \textit{Target Selection Bottleneck}: without a clear target, perturbation-based explanations cannot be effectively anchored. To support this (addressing RQ2), we propose moving beyond manual specification. By employing KD-Tree partitioning \cite{bentley1975multidimensional}, our framework proactively segments the objective landscape, offering local ``dominating points'' as natural, data-driven targets for explanation.

\subsection{Distance-Based Learning and Perturbation Analysis}

The third avenue of research (RQ3) addresses the mechanics of generating geometrically meaningful explanations. Although standard perturbation methods such as SHAP \cite{lundberg2017unified} and LIME \cite{ribeiro2016should} are powerful for explaining scalar predictions or classification probabilities, they do not inherently model the \textit{spatial displacement} of a solution within a multi- or many-objective landscape. To capture this geometric sensitivity, we draw upon the \textit{Distance Explainer} methodology proposed by Meijer and Bos \cite{meijer2025explainable} to interpret deep learning embeddings.

A critical component of this methodology is the adaptation of the Randomized Input Sampling for Explanation (RISE) framework \cite{petsiuk2018rise}. Originally designed for image saliency, RISE estimates feature importance by probing a black-box model with random binary masks and aggregating the resulting degradation in the model's output. In our context, we reframe this to measure the ``shift in distance'': we define saliency as the degree to which masking a decision variable causes a solution to drift away from a target in the objective space.

However, applying RISE-based perturbations directly to expensive evaluation functions is computationally prohibitive. This necessitates a surrogate model that is not only fast but also explicitly preserves the geometric structure of the data. The Minimal Learning Machine (MLM) \cite{deSouza2013mlm} serves as a bridge. Unlike standard neural networks that act as opaque mappings, MLM constructs a regression model based entirely on distance matrices. This unique property creates a direct mathematical alignment with the Distance Explainer, allowing us to rigorously quantify how perturbations in the decision space translate into movement in the objective space, thereby providing the robust foundation required for the Partition-Guided Distance Saliency framework.

\section{Methodology: The Partition-Guided Distance Saliency (PGDS) Framework}\label{sec:methodology}

The proposed framework enables a continuous, geometry-aware explainability pipeline for many-objective optimization. It operates in three distinct stages: (1) training a distance-preserving surrogate model (MLM) to map decision space geometry to objective space; (2) partitioning the objective landscape via KD-Trees to identify automated ``Regions of Interest'' and their corresponding dominating targets; and (3) generating saliency maps via tabular perturbation to quantify the influence of decision variables on a solution's proximity to these targets.

\subsection{ Geometric Surrogate Modeling via Minimal Learning Machines}
To explain the geometric displacement of a solution, we require a surrogate model that predicts distances rather than scalar objective values. We employ the Minimal Learning Machine (MLM) \cite{deSouza2013mlm}, a supervised learning method that learns a mapping between input and output distance matrices.

Let $\mathcal{X} \in \mathbb{R}^n$ be the decision space and $\mathcal{Y} \in \mathbb{R}^M$ be the objective space. Given an archive of $N$ solutions, we randomly select a subset of $K$ reference points from this set. As established in the original MLM formulation \cite{deSouza2013mlm}, the set of reference points is drawn from the training data itself, and the parameter $K$ serves as a critical hyperparameter. The method's ability to accurately reconstruct the geometric manifold is sensitive to this choice; typically, $K$ is set as a function of $N$ (e.g., $K = \lceil \sqrt{N} \rceil$) to balance the regression error with the computational complexity of the subsequent matrix inversion.

Considering the construction of the distance matrix, we compute the input distance matrix $D_x \in \mathbb{R}^{N \times K}$ and the output distance matrix $\Delta_y \in \mathbb{R}^{N \times K}$, in which each entry represents the Euclidean distance between a data point and a reference point:
\begin{equation}
D_{x(i,k)} = ||\mathbf{x}_i - \mathbf{r}_k||_2, \quad \Delta_{y(i,k)} = ||\mathbf{y}_i - \mathbf{t}_k||_2,
\end{equation}
in which $\mathbf{r}_k$ and $\mathbf{t}_k$ are the $k$-th reference points in the decision and the objective space, respectively.

The MLM assumes a linear mapping between these geometries:
\begin{equation}
\Delta_y = D_x \mathbf{B} + \mathbf{E}.
\end{equation}
The regression matrix $\mathbf{B} \in \mathbb{R}^{K \times K}$ is estimated via Ordinary Least Squares (OLS), providing a closed-form solution with complexity $\Theta(K^2N)$:
\begin{equation}
\hat{\mathbf{B}} = (D_x^T D_x)^{-1} D_x^T \Delta_y.
\end{equation}
For a new query solution $\mathbf{x}_{query}$, we compute its input distance vector $\mathbf{d}_{in} = [||\mathbf{x}_{query} - \mathbf{r}_1||, \dots, ||\mathbf{x}_{query} - \mathbf{r}_K||]$. The predicted output distances are given by $\hat{\boldsymbol{\delta}}_{out} = \mathbf{d}_{in} \hat{\mathbf{B}}$. Finally, the coordinates in the objective space $\hat{\mathbf{y}}$ are reconstructed from $\hat{\boldsymbol{\delta}}_{out}$. This process is intuitively similar to the operation of a Global Positioning System (GPS). In GPS, a receiver determines its location not by directly measuring the coordinates, but by knowing its distance to several satellites (reference points) and finding the intersection of the spheres centered at those satellites.

Similarly, the MLM has predicted how far the target solution should be from each of the $K$ reference points in the objective space. We must therefore find the coordinate vector $\hat{\mathbf{y}} \in \mathbb{R}^M$ that best satisfies these $K$ distance constraints simultaneously. This is formulated as a non-linear optimization problem where we seek to minimize the discrepancy between the predicted distances and the actual Euclidean distances to the reference points:
\begin{equation}
\hat{\mathbf{y}} = \arg\min_{\mathbf{y} \in \mathbb{R}^M} \sum_{k=1}^{K} \left( ||\mathbf{y} - \mathbf{t}_k||_2 - \hat{\delta}_{out, k} \right)^2.
\end{equation}
This multilateration problem is typically solved using the Levenberg-Marquardt algorithm \cite{deSouza2013mlm}, which iteratively adjusts the coordinates to triangulate the solution's precise position in the objective landscape.

The selection of reference points $\mathcal{R}$ introduces a stochastic component to the surrogate training. However, the MLM is globally robust to this selection; provided $K$ is sufficiently large to capture the archive's topology, the resulting saliency rankings remain stable across different random seeds \cite{deSouza2013mlm}.

Additionally,  the MLM assumes a linear mapping between distance matrices; this does not imply a linear relationship between the decision variables and the objectives. The choice to map distance matrices, ${D}_{x}$ and  ${\Delta}_{y}$, rather than raw coordinate vectors, stems from the need to preserve the structure manifold of the Pareto front. By modeling the relationship between distances, the MLM sidesteps the complexities of high-dimensional coordinate transformations, which are often non-linear and ill-conditioned in Many-Objective landscapes. The regression matrix $\bm{B}$ acts as a linear geometric bridge. It assumes that if two points are close in the decision space, their relative distances to a set of global ``anchors'' (reference points) should transform linearly into the objective space. This allows the model to simultaneously predict the proximity of a query point to the entire archive, rather than predicting a single point-to-point coordinate.

\subsection{Automated Target Discovery via KD-Tree Partitioning}
In many-objective scenarios, the Decision Maker (DM) often lacks the intuition to manually specify a target point $T$. To automate this, we employ K-Dimensional Tree (KD-Tree) partitioning \cite{bentley1975multidimensional} to segment the objective space into hyper-rectangular blocks.

The archive $\mathcal{Y}$ is recursively split along the dimension with the highest variance. The process terminates when a maximum depth $d_{max}$ is reached, or a block contains fewer than $N_{min}$ points. This creates a set of leaf nodes $\{L_1, \dots, L_B\}$.

For each leaf node $L_j$, we define a local \textit{Dominating Point} $P_{dom}^{(j)}$, which serves as the automated target for that region. Assuming without loss of generality that all objectives are formulated for minimization, this point is constructed from the minimum observed coordinates within the block boundaries:
\begin{equation}
P_{dom}^{(j)} = [\min_{y \in L_j}(f_1), \min_{y \in L_j}(f_2), \dots, \min_{y \in L_j}(f_M)].
\end{equation}

This point represents the ``local utopia'', the best possible trade-off available within that specific region of the objective space. Importantly, these points serve as automated suggestions to facilitate exploration. The framework presents these targets alongside the explicit partition rules defining their respective regions (e.g., $Region_1: f_1 < 0.5 \land f_2 > 0.8$). This empowers the DM to inspect the logical boundaries of each block and proactively disregard regions or targets that do not align with their specific preferences or domain constraints.

\subsection{Saliency maps via Distance Explainers}\label{section saliency maps}
With a trained surrogate and a defined target $T$ (either manually selected or automatically set to $P_{dom}^{(j)}$), we quantify the variable importance using the Distance Explainer principle \cite{meijer2025explainable}. We adapt the RISE framework \cite{petsiuk2018rise} to generate tabular perturbations.

Let $m \in \{0, 1\}^n$ be a binary mask vector in which $m_i=0$ indicates a ``masked'' feature. We define a baseline reference $\bar{x}$, representing the global mean of the archive. The perturbed sample $\tilde{x}$ is generated as:
\begin{equation}
\tilde{x} = x_{query} \odot m + \bar{x} \odot (1 - m).
\end{equation}
This operation effectively ``neutralizes'' the contribution of masked variables by reverting them to the population mean.

Considering saliency calculation, we generate $L$ random masks. For each mask $\mathbf{m}^{(l)}$, we compute the perturbed output distance to the target $T$:
\begin{equation}
d_{shift}^{(l)} = ||\text{MLM}(\tilde{x}^{(l)}) - T||_2.
\end{equation}

The saliency score $S_i$ for the decision variable $x_i$ is calculated as the correlation between the variable's presence in the mask and its resulting proximity to the target.  A high score indicates that when $x_i$ is preserved (unmasked), the solution stays close to the target; when masked, it drifts away.

The final saliency map $S$ is derived by correlating the presence of each variable with the resulting distance shifts. We distinguish between two types of influence:
\begin{itemize}
    \item \textbf{Drivers}: Variables that, when preserved (unmasked), significantly \textit{reduce} the distance to the target. These are the levers the DM must pull to converge toward the region's dominating point.
    \item \textbf{Blockers}: Variables that, when preserved, \textit{increase} or maintain a large distance from the target. These indicate decision values that are currently hindering the solution from entering the optimal zone of the selected region.
\end{itemize}

\subsection{Algorithm Summary}\label{section Algorithm Summary}
The complete procedure for the Partition-Guided Distance Saliency (PGDS) framework is detailed in Algorithm \ref{alg:pgds}. 

The algorithm accepts the final optimization archive, comprising decision vectors $X$ and objective vectors $Y$, along with the specific query solution $x_q$ to be explained. Three hyperparameters control the process\footnote{The hyperparameters $K$ (anchors) and $L$ (sampling) represent a trade-off between resolution and cost. $K$ determines the rigidity of the geometric surrogate model, while $L$ ensures the statistical convergence of the saliency scores. In our experiments, as will be seen in Section \ref{sec:experiments}, these were tuned to ensure $R^2>0.99$, ensuring that the generated explanations are anchored in a high-fidelity geometric mapping. \\ Additionally, the KD-Tree depth D is a user-specified parameter that dictates the granularity of the objective space partitioning. This allows the DM to adapt the ``Regions of Interest'' to their specific exploration goals and the desired solution density within each partitioned block.} $K$ defines the number of geometric anchors for the surrogate; $D$ controls the granularity of the KD-Tree partition; and $L$ sets the number of random masks used for sampling. The procedure returns two key artifacts: the automated \textbf{target} $T$ (the geometric point of reference) and the \textbf{saliency map} $S$, which quantifies the influence of each decision variable on the solution's proximity to that target.

\begin{algorithm}
\caption{Partition-Guided Distance Saliency (PGDS)}
\label{alg:pgds}
\begin{algorithmic}[1]
\State \textbf{Input:} Archive $(X, Y)$, Query $x_q$, max depth $D$, References $K$, Masks $L$.
\State \textbf{Step 1: Surrogate Training (MLM)}
\State Select set of $K$ reference points $\mathcal{R} = \{\mathbf{r}_1, \dots, \mathbf{r}_K\}$ randomly from Archive $X$
\State Compute Distance Matrices $D_x$ and $\Delta_y$
\State Solve for Regression Matrix: $\hat{B} = (D_x^T D_x)^{-1} D_x^T \Delta_y$
\State \textbf{Step 2: Automated Target Identification}
\State Build KD-Tree on $Y$ with depth $D$
\State Identify leaf block $L_j$ containing $x_q$
\State Set Target $T \leftarrow P_{dom}^{(j)}$ (min coords of $L_j$)
\State \textbf{Step 3: Saliency Map Generation (Distance Explainer)}
\For{$l = 1$ to $L$}
    \State Generate random binary mask $m^{(l)}$
    \State Perturb Input: $\tilde{x} \leftarrow x_q \odot m^{(l)} + \bar{x} \odot (1-m^{(l)})$
    \State \textbf{Predict Position:}
    \State \quad Compute input distances: $d_{in} \leftarrow ||\tilde{x} - \mathcal{R}||$
    \State \quad Predict output distances: $\hat{\delta}_{out} \leftarrow d_{in}\hat{B}$
    \State \quad Multilaterate: $\hat{y} \leftarrow \min_y \sum (||\hat{y} - T_{refs}|| - \hat{\delta}_{out})^2$
    \State Compute Distance Shift: $dist^{(l)} \leftarrow ||\hat{y} - T||$
\EndFor
\State Compute Saliency $S$ via correlation of $m$ and $dist$
\State \textbf{Return:} Saliency Map $S$, Target $T$
\end{algorithmic}
\end{algorithm}

The process begins with the surrogate training step (Lines 2-5). Here, the Minimal Learning Machine is initialized by randomly selecting $K$ reference points from the archive $X$ to serve as geometric anchors. The input and output distance matrices ($D_x$ and $\Delta_y$) are calculated, and the regression matrix $\hat{B}$ is solved via Ordinary Least Squares (Line 5), creating the global topological mapping.

The second step constitutes the automated target identification (Lines 6-9), addressing the decision-making bottleneck. A KD-Tree is constructed on the objective vectors $Y$ to partition the space (Line 7). For a given query solution $x_q$, the algorithm identifies the specific leaf block $L_j$ containing it and automatically sets the target $T$ to be the ``Dominating Point'' ($P_{dom}^{(j)}$) of that region (Line 9), defined by the minimum coordinates within the block boundaries.

The final step represents the saliency map generation (Lines 10-18), which adapts the RISE framework. The algorithm iterates through $L$ randomly generated binary masks (Lines 11-12), perturbing the query solution by replacing masked features with the population mean $\bar{x}$ (Line 13). 

For each perturbed sample, the position is predicted via the MLM surrogate chain: calculating input distances, projecting them to output distances, and reconstructing the coordinates via multilateration (Lines 14-17). 

Finally, the shift in distance relative to the target $T$ is recorded (Line 18), and the variable saliency is computed as the correlation between the mask presence and the proximity to the target (Line 20).

An implemented version in Python is available at \href{https://github.com/ClaudioLucioLopes/PGDS}{PGDS}.

\section{Experiments}\label{sec:experiments}

\subsection{Visualizing the PGDS Framework: A Step-by-Step Walkthrough}

\begin{figure}[htbp]
    \centering
     \adjustbox{trim=4cm 0.3cm 4cm 0.3cm}{%
    \includesvg[width=0.8\textwidth]{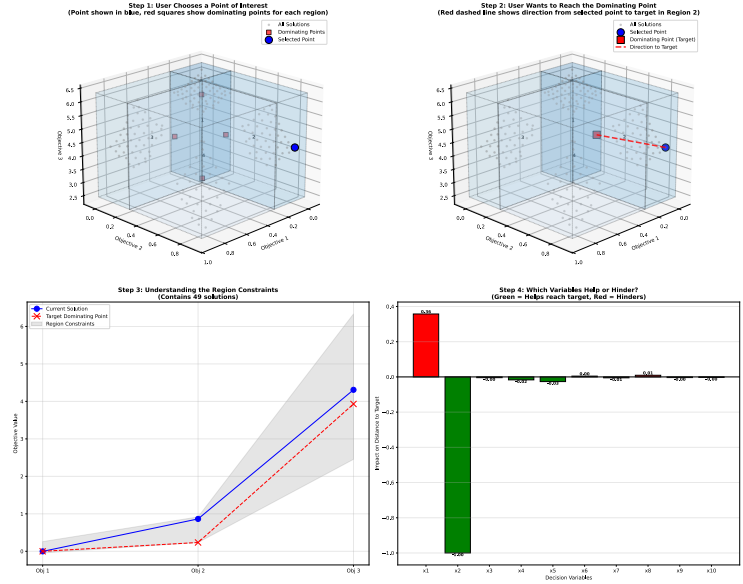}
    }
    \caption{{The user journey through PGDS.} \textit{Top-left}: the space is partitioned into cuboids, offering automated Dominating Points (red squares). \textit{Top-tight}: a trajectory is defined from the user's selection to the local target. \textit{Bottom-left}: the constraints of the region are visualized. \textit{Bottom-right}: the saliency map reveals that variable $x_2$ is the primary driver for reaching the target in this specific region.}
    \label{fig:pgds_walkthrough}
\end{figure}

To demonstrate the practical utility of the PGDS framework, we present a walkthrough using the DTLZ7 benchmark problem \cite{Huband2011}. DTLZ7 is chosen for its disconnected, multi-modal Pareto front, which presents a significant challenge for traditional visualization. The process is illustrated in Figure \ref{fig:pgds_walkthrough}, which maps the algorithmic steps from Section \ref{section Algorithm Summary} to the DM's perspective.

\begin{figure}[htbp]
    \centering
    \adjustbox{trim=3cm 1cm 3cm 0cm}{%
    \includesvg[width=0.8\textwidth]{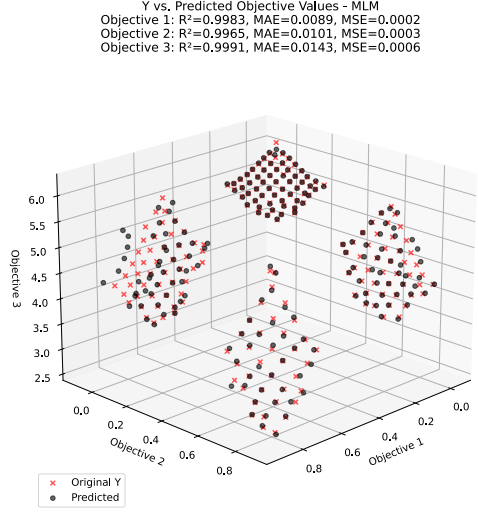}
    }
    \caption{{Surrogate model validation on DTLZ7.} The alignment between original data (red crosses) and MLM predictions  (black dots) demonstrates the model's ability to preserve the complex, disconnected geometry of the Pareto front with high precision ($R^2 \approx 0.99$).}
    \label{fig:mlm_fit}
\end{figure}

Before any explanation can be generated, the system must learn to predict the problem. As detailed in Algorithm \ref{alg:pgds}, this process relies on the hyperparameter $K$ (Reference Points), which determines the number of geometric anchors randomly sampled from the archive. In this specific validation, $K$ was set to half the total number of solutions ($K = N/2$). As shown in Figure \ref{fig:mlm_fit}, the MLM model is trained on the optimization archive. The plot overlays the original objective vectors, represented by red crosses, with the MLM's predictions represented by black dots. The high coefficient of determination, $R^2 > 0.99$, and negligible Mean Squared Error, MSE $\approx 0.0002$, across all three objectives confirm that the surrogate has successfully captured the manifold's geometry. This high-fidelity mapping is the prerequisite for reliable distance-based explanations.

Once the model is ready, the interaction begins, Figure \ref{fig:pgds_walkthrough}, \textit{top-left}. The KD-Tree algorithm partitions the disconnected search space into distinct hyper-rectangular blocks (blue transparent cuboids). The system automatically identifies a ``Dominating Point'' (small red square) for each region. Instead of facing a ``cognitive drought'' and staring at a cloud of points, the DM is presented with structured options. The DM simply selects a solution of interest, the blue circle, within one of these regions.

The framework then contextualizes the selected solution, Figure \ref{fig:pgds_walkthrough}, \textit{top-right}. It draws a geometric vector, represented by the red dashed line, connecting the user's selection, given by the blue circle, to the region's automated dominating point, represented by the red square. This vector defines the ``direction of improvement'' specific to that local trade-off region. It answers the implicit question: ``What is the best I can theoretically achieve if I stay within this specific configuration of constraints?''

A key advantage of the PGDS framework is the ability to visualize the ``why'' behind a region. The \textit{Region constraints plot}, Figure \ref{fig:pgds_walkthrough}, \textit{bottom-left}, directly visualizes the boundary rules of the leaf block $L_j$ identified in Algorithm \ref{alg:pgds} (Line 8). \textit{Gray Area, the rules:} This shaded band represents the min/max bounds of the current region ($L_j$) for every objective. It answers the question: ``What defines this region?''. \textit{Trajectory:} The plot shows the gap between the User's selection (blue circle) and the region's dominating point (red cross). This allows the DM to instantly see which specific objective is hitting the region's boundary.

Finally, the system explains \textit{how} to traverse the trajectory toward the target, Figure \ref{fig:pgds_walkthrough}, \textit{bottom-right}. The Saliency Map explicitly categorizes the decision variables based on the correlation logic defined in Section \ref{section saliency maps}:
\begin{itemize}
    \item {Green bars, the \textbf{Drivers}:} Variables like $x_2$ show a strong negative impact on distance, meaning they are the primary levers to pull to move the solution closer to the target.
    \item {Red bars, the \textbf{Blockers}:} Variables like $x_1$ show a positive impact, indicating that their current values or perturbations might be hindering progress toward the target.  
\end{itemize}

This transforms the abstract geometric distance into actionable engineering insight: ``To reach improvements in this region using this target, adjust $x_1$ to remove the bottleneck while carefully maintaining the driver $x_2$.''
     
\begin{figure}[htbp]
    \centering

    \includesvg[width=0.8\textwidth]{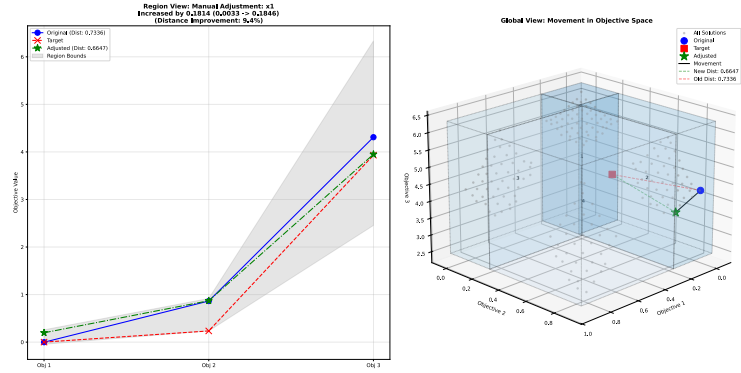}
    \caption{{Validating the ``Blocker'' Hypothesis.} Variable $x_1$ was identified as a blocker (red bar in saliency map). Manually increasing $x_1$ results in a $9.4\%$ improvement in the solution's proximity to the target (Global View, green vector), confirming that the variable was indeed hindering convergence.}
    
    \label{fig:pgds_walkthrough_blocker}
\end{figure}

To verify the actionable insights generated by the Saliency Map, we perform a sensitivity analysis by aligning the identified high-saliency variables with the configuration of the region's Dominating Point. Instead of arbitrary tuning, we calculate the precise delta ($\Delta_i = x_{target, i} - x_{current, i}$) required to shift the variable from its current state to the optimal value observed in the target solution. We then apply this calculated shift and observe the resulting impact on the geometric distance in the objective space.

First, we address the identified \textbf{Blocker}, variable $x_1$ (Figure \ref{fig:pgds_walkthrough_blocker}). The system calculates that $x_1$ must be increased by $0.18$ to match the target configuration. Applying this specific adjustment results in a $9.4\%$ improvement (decrease) in the Euclidean distance to the target, confirming that the variable indeed hindered convergence.

\begin{figure}[htbp]
    \centering
    \includesvg[width=0.8\textwidth]{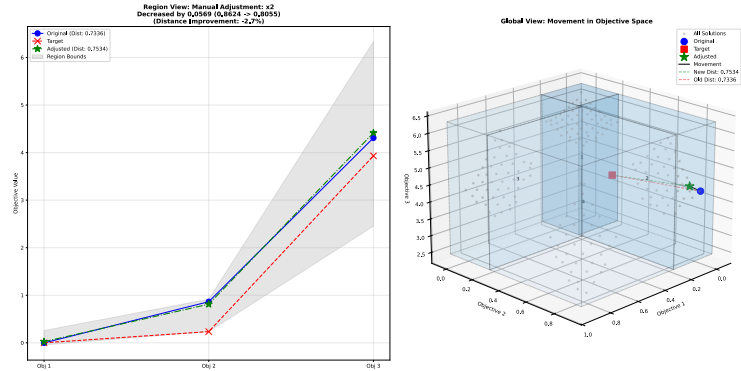}
    \caption{{Validating the ``Driver'' Hypothesis.} Variable $x_2$ was identified as a driver (green bar). Decreasing $x_2$ leads to a $2.7\%$ degradation (distance increase) in performance. This sensitivity confirms that $x_2$ is a critical component of the solution's success and must be carefully managed.}
    \label{fig:pgds_walkthrough_driver}
\end{figure}

Second, we test the identified \textbf{Driver}, variable $x_2$ (Figure \ref{fig:pgds_walkthrough_driver}). The system identifies that $x_2$ is critical. When we deviate from the calculated target trajectory by decreasing $x_2$ by $0.05$, the distance to the target worsens by $2.7\%$. This degradation confirms that $x_2$ plays a critical role in maintaining proximity to the optimal front and must be preserved.

\subsection{Physics-Informed Validation (Welded Beam)}
\label{sec:exp_welded_beam}

The second experiment addresses RQ3, regarding sensitivity, and RQ1, interpretability, by applying PGDS to the Welded Beam Design problem \cite{deb_sundar_2006}. This problem is chosen because the relationships among decision variables (weld thickness $h$, length $l$, height $t$, width $b$) and constraints (shear stress $\tau$, bending stress $\sigma$, buckling load $P_c$) are governed by known physical laws. This allows us to verify whether the mathematically derived blockers correspond to actual physical limitations or not.

The engineering objective is to design a welded beam that minimizes the fabrication cost ($f_1$) and the end deflection ($f_2$). The problem consists of four decision variables $\vec{x} = (h, l, t, b)$ and is subject to four non-linear constraints, as formulated in \cite{deb_sundar_2006}:
\begin{align}
\text{Minimize } & f_1(\vec{x}) = 1.10471h^2l + 0.04811tb(14.0+l), \\
\text{Minimize } & f_2(\vec{x}) = \frac{2.1952}{t^3b},
\end{align}
subject to:
\begin{align}
g_1(\vec{x}) & \equiv 13,600 - \tau(\vec{x}) \ge 0, \quad (\text{Shear Stress}) \\
g_2(\vec{x}) & \equiv 30,000 - \sigma(\vec{x}) \ge 0, \quad (\text{Bending Stress}) \\
g_3(\vec{x}) & \equiv b - h \ge 0, \quad (\text{Weld Geometry}) \\
g_4(\vec{x}) & \equiv P_c(\vec{x}) - 6,000 \ge 0, \quad (\text{Buckling Load})\\
0.125 & \leq b, h \leq 5.0, \quad  \\
0.1 & \leq l,t \leq 10.0. \quad  
\end{align}

For brevity, we present the high-level constraint definitions above. The detailed constitutive equations governing the physical mechanics
involve complex interactions between the beam geometry and applied forces. The complete mathematical derivations and parameter constants are fully detailed in \cite{deb_sundar_2006}.

To test the framework's sensitivity, we utilized NSGA-II \cite{deb_nsgaII_2002} to generate an approximation of the Pareto front (200 solutions) and selected a specific ``Constraint Trap'' solution.

\begin{figure}[htbp]
    \centering
    \includesvg[width=0.9\textwidth]{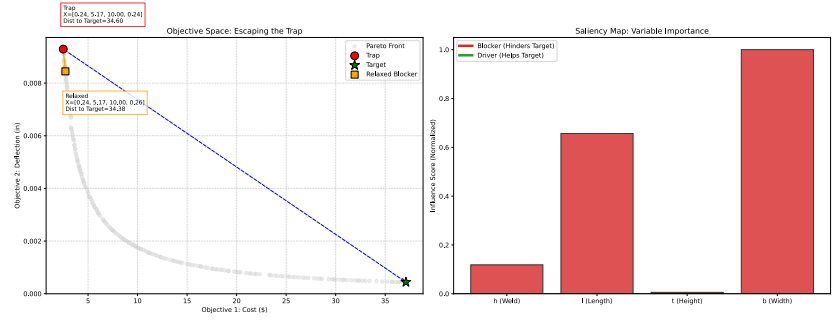}
    \caption{{Physics-Informed Validation on Welded Beam.} Left plot is the ``Trap Solution'' (red Circle), that is stuck at a low-cost optimum. PGDS identifies that to reach the high-performance target (green star), the solution must escape the buckling constraint. Right plot presents the saliency map, which identifies beam width ($b$) as the primary ``Blocker'' (highest red bar), correctly signaling that width is one of the physical bottlenecks preventing improvement.}
    \label{fig:welded_beam}
\end{figure}

While Algorithm \ref{alg:pgds} provides an automated target, the PGDS framework also
supports manual target selection. As detailed in Figure \ref{fig:welded_beam}, left plot, the selected solution (red circle) is located in the low-cost region of the Pareto front ($cost \approx 2.50$). However, this efficiency comes at a price: the solution is physically precarious. An analysis of its constraint values reveals it is positioned almost exactly on the boundary of the buckling constraint ($g_4 \approx -0.0001$), meaning the beam is on the verge of structural failure.

We executed the PGDS with a distinct goal: rather than a local dominating point, we set the target to the best deflection point found in the archive (green star in Figure \ref{fig:welded_beam}). This evaluates whether the method can steer a low-cost design towards a high-performance configuration. 

The resulting saliency map (Figure \ref{fig:welded_beam}, right plot, identifies the variable {$b$ (Width)} as the overwhelming ``Blocker'' with a normalized saliency value of $1.0$.
While other variables length $l$ show some blocking influence, the framework signals that $b$ is the critical variable preventing the solution from moving toward the target.

To validate this diagnosis, we performed a sensitivity analysis by perturbing the identified blocker {$b$ (Width)}. As shown in Table \ref{tab:welded_verification}, blindly adjusting other variables while keeping $b$ fixed at its trap value ($0.236$) would likely result in an infeasible design due to the active buckling constraint. However, relaxing $b$ (increasing it to $0.259$) immediately relieves the pressure on the buckling constraint, shifting $g_3$ from a critical $-0.0001$ to a safe $-0.0049$.

This confirms that PGDS successfully detected the ``wall'' created by the buckling equation $P_c(x)$ without explicit knowledge of the formula. By flagging {$b$ (Width)} as a blocker, the framework correctly informed the DM that no significant improvement in deflection is possible without first increasing the beam's width.

\begin{table}[htbp]
    \centering
    \caption{Physics-Informed Verification of the ``Blocker'' Hypothesis}
    \label{tab:welded_verification}
    \begin{tabular}{l|c|c|c}
        \hline
        \textbf{Component} & \textbf{Trap Value} & \textbf{Relaxed Value} & \textbf{Physical Impact} \\
        \hline
        \textbf{Variable $b$ (Width)} & \textbf{0.2363} & \textbf{0.2599} & \textbf{Blocker Relaxed} \\
        Variable $h$ (Weld) & 0.2358 & 0.2358 & Unchanged \\
        Variable $l$ (Length) & 5.1674 & 5.1674 & Unchanged \\
        \hline
        \textbf{Buckling ($g_3$)} & \textbf{-0.0001} & \textbf{-0.0049} & \textbf{Constraint Satisfied} \\
        Shear Stress ($g_1$) & -0.0016 & -0.0016 & Stable \\
        \hline
    \end{tabular}
\end{table}

\subsection{Experiment 3: Scalability to Many-Objective Optimization (WFG3)}
\label{sec:exp_wfg3}

The final experiment examines RQ2, concerning automated discovery, and RQ1, concerning interpretability, in the setting of many-objective optimization. As the number of objectives increases beyond three, traditional scatter plots become unintelligible, making it difficult for the DM to visually identify interesting trade-offs. We utilize the PGDS framework to help navigate and explain a 10-objective landscape.

We selected the WFG3 \cite{Huband2011} benchmark problem configured with 10 objectives and 20 decision variables. WFG3 is particularly challenging because it features a degenerate Pareto front (a lower-dimensional curve embedded in high-dimensional hyperspace), making it an ideal test for the geometric sensitivity of the PGDS.

The experimental protocol proceeded as follows:
\begin{enumerate}
    \item {Optimization:} We employed NSGA-III to generate an approximation of the Pareto front. The algorithm ran for 400 generations, producing a final archive of 108 non-dominated solutions.
    \item {Blind Targeting:} Unlike previous experiments where targets were visually verified, here we relied entirely on the KD-Tree to partition the 10-dimensional objective space into semantic blocks.
    \item {Region Selection Logic}: While PGDS automatically partitions the space into hyper-rectangular semantic blocks, we applied standard decision-making heuristics to select specific regions for explanation. This simulates a DM prioritizing regions based on common goals:
    \begin{itemize}
        \item The \textbf{Knee Region}: The block containing the solution with the minimum Euclidean distance to the ideal point (the origin).
        \item The \textbf{Extreme Region}: The block containing the solution with the best value for objective $f_{10}$, regardless of other trade-offs.
    \end{itemize}
\end{enumerate}

\begin{figure}[htbp]
    \centering
    \includesvg[width=0.8\textwidth]{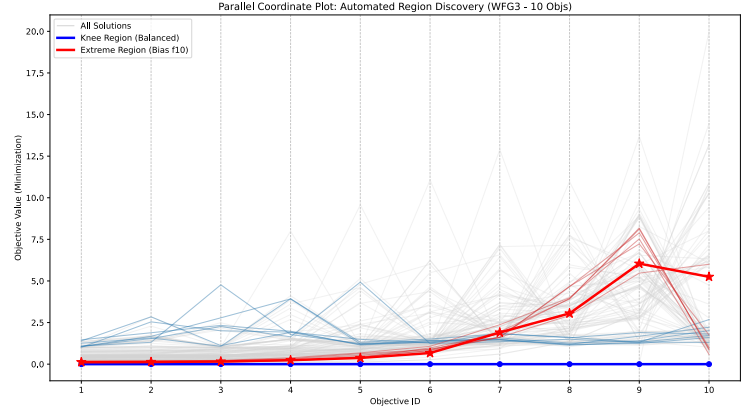}
    \caption{Region selection discovery in 10 Dimensions. The Parallel Coordinate Plot reveals the contrasting geometries identified by the KD-Tree. The \textbf{knee} region (Blue) represents balanced solutions, while the \textbf{extreme} region (Red) captures solutions optimizing specific objectives ($f_{10}$) at the expense of others ($f_9$). The bold lines are the dominating solutions in each region.}
    \label{fig:wfg3_pcp}
\end{figure}

The PGDS framework successfully distinguished the two regions without human intervention. Figure \ref{fig:wfg3_pcp} presents a Parallel Coordinate Plot (PCP) visualizing the Knee Region (Blue), which exhibits a balanced profile across all 10 objectives, whereas the Extreme Region (Red) shows aggressive trade-offs.

\begin{figure}[htbp]
    \centering
    \includesvg[width=0.8\textwidth]{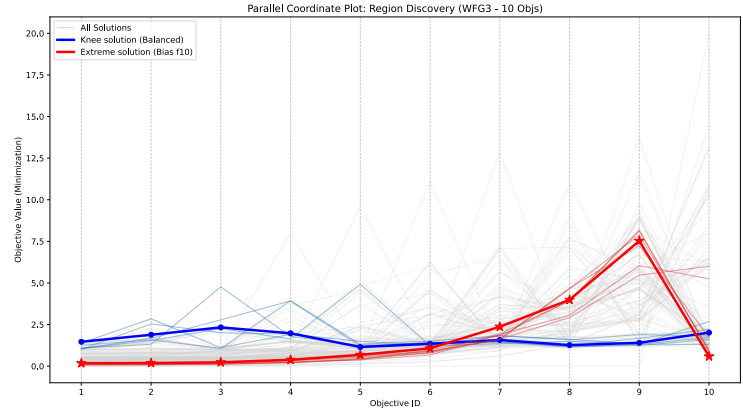}
    \caption{Representative solutions selected for saliency analysis in WFG3-10. Unlike Figure \ref{fig:wfg3_pcp}, which highlights the local utopian boundaries (dominating points) of the KD-Tree partitions, this plot visualizes the actual solutions identified in Table \ref{tab:wfg3_targets}, the balanced Knee Point (blue bold) and the Extreme Point (red bold) optimized for $f_{10}$.}
    \label{fig:wfg3_pcp_sol}
\end{figure}

To resolve the ``cognitive drought'' inherent in many-objective visualization, we distinguish between the \textit{geometrical potential} of a region and the \textit{actual solution} analyzed. Figure \ref{fig:wfg3_pcp} visualizes the ``Dominating Points'' (local utopias) that define the boundaries of each partition. In contrast, Figure \ref{fig:wfg3_pcp_sol} illustrates the specific representative solutions from Table \ref{tab:wfg3_targets}. This distinction explains the trade-off profile: the bold red solution in Figure \ref{fig:wfg3_pcp_sol} achieves a superior, lower value at $f_{10}$ compared to the knee solution, which sacrifices $f_{10}$  to maintain global balance across the other nine objectives.

\begin{table}[htbp]
    \centering
    \caption{Automated ``Blind'' Targets identified by PGDS in WFG3}
    \label{tab:wfg3_targets}
    \resizebox{\textwidth}{!}{
    \begin{tabular}{l|c|c|l}
        \hline
        \textbf{Region Type} & \textbf{Key Objective Values} & \textbf{Dist. to Ideal} & \textbf{Selection Logic} \\
        \hline
        \textbf{Knee Point} & $f_1=1.46, \dots, f_{10}=2.01$ & $5.32$ & Balanced trade-off (min dist to origin). \\
        \textbf{Extreme Point} & $f_1=0.17, f_9=7.52, \mathbf{f_{10}=0.58}$ & N/A & Best possible value for $f_{10}$. \\
        \hline
    \end{tabular}
    }
\end{table}

Table \ref{tab:wfg3_targets} details the specific automated targets found. Note that while the Knee point maintains a middle ground (all objectives $\approx 1.5 - 2.0$), the Extreme point achieves a superior $f_{10}$ ($0.58$) but sacrifices $f_9$ ($7.52$).

\begin{figure}[htbp]
    \centering
    \includesvg[width=1\textwidth]{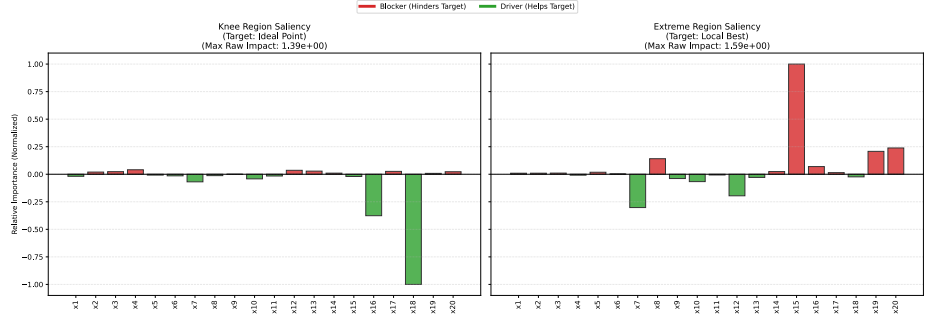}
    \caption{{Context-aware saliency comparison.}  In the Knee Region, the left one, variable $x_{18}$ is the primary {driver} (green) facilitating convergence. In the right plot, the extreme region of the variable $x_{15}$ becomes a dominant \textbf{Blocker} (Red), hindering performance. This proves that variable importance is not static but dependent on the region of the Pareto front.}
    \label{fig:wfg3_saliency}
\end{figure}

The most critical finding using the saliency analysis is the explanation, illustrated in Figure \ref{fig:wfg3_saliency}. \textbf{Knee Region}: in the left plot, maintaining the balanced knee position, the system identifies the variable ${x_{18}}$ as the primary {driver} (Score: $-1.00$). The variable $x_{16}$ also plays a supporting driver role. This suggests that $x_{18}$ is the keystone variable for global balance in WFG3. 
\textbf{Extreme Region}: In the right plot, the landscape changes completely. Here, the variable ${x_{15}}$ emerges as a massive blocker (score: $+1.00$). This indicates that $x_{15}$ is the distinct ``constraint'' preventing further optimization of $f_{10}$ or causing the severe trade-off with $f_9$.

The identified saliency for $x_{18}$ (Knee) and $x_{15}$ (Extreme) provides actionable insights rooted in the benchmark's structure. In WFG3, these specific variables act as keystone position parameters that control the underlying manifold's curvature. By flagging $x_{15}$ as a ``Blocker'' in the Extreme region, PGDS correctly identifies the specific decision variable whose current configuration prevents the solution from sliding further down the $f_{10}$ boundary without degrading other objectives.

The experiment confirms that PGDS scales effectively to high-dimensional problems. Even without visual aids, the framework provided differentiated, actionable insights: telling the DM to focus on $x_{18}$ for a balanced solution, but to manipulate $x_{15}$ if minimizing $f_{10}$ is the priority. This capability effectively bridges the interpretability gap in Many-Objective Optimization.

In many-objective optimization, traditional visualization represents a ``failing baseline'' in decision and objective explainability, because a standard PCP, the gray lines in Figures \ref{fig:wfg3_pcp} and \ref{fig:wfg3_pcp_sol}, for example, provides no guidance on where to look or which variables to adjust. PGDS overcomes this by overlaying automated targets and providing the saliency map, as in Figure \ref{fig:wfg3_saliency}, which transforms the intelligible set of solutions into a directed set of engineering instructions—identifying exactly which levers ($x_{18}$) to pull for balance.

\section{Conclusion}\label{sec:conclusion}
\label{sec:conclusion}

This work introduced the {Partition-Guided Distance Saliency (PGDS)} framework, a methodology engineered to bridge the interpretability gap between high-dimensional decision spaces ($\mathcal{X}$) and complex, many-objective landscapes ($\mathcal{Y}$). By synergizing the geometric learning capabilities of Minimal Learning Machines (MLM) with the automated spatial decomposition of KD-Trees, PGDS effectively resolves the ``cognitive drought'' often experienced by DMs when traditional visualization techniques fail.

Our experimental validation confirms the framework's robustness across some perspectives: convergence, diversity, and scalability. Collectively, these results establish PGDS as a diagnostic instrument. It empowers DMs to navigate the Pareto front without \textit{a priori} target knowledge, automating the identification of ``Dominating Points'' within local regions and rigorously quantifying the directional influence of decision variables through distance-based saliency.

A direct quantitative comparison with existing rule-based XAI methods (such as XLEMOO or R-XIMO) remains due to the differing nature of their outputs—logical predicates versus continuous saliency maps. Future research should focus on developing standardized explanation quality indicators to benchmark distance-based saliency against classification-based attribution in high-dimensional spaces.

A systematic sensitivity analysis of the $K$, $D$, and $L$ hyperparameters is required to rigorously define the bounds of the framework’s robustness and provide practitioners with automated tuning heuristics for varying archive sizes.

While PGDS successfully classifies variables as \textit{Drivers} or \textit{Blockers} based on their geometric influence, the current iteration operates as a {diagnostic} rather than a {prescriptive} framework. The methodology relies on sensitivity analysis to estimate the impact of perturbations; however, it does not yet mathematically invert the surrogate model to yield the precise decision vector $\mathbf{x}^*$ required to reach a specific target $T$. This limitation stems from the inherent difficulty of the inverse problem, where the mapping from a high-dimensional decision space ($\mathbb{R}^n$) to another dimensional objective space ($\mathbb{R}^m$) can be ill-posed and involve significant information loss.

Another future research will focus on evolving PGDS from a descriptive tool into a fully prescriptive engine. Our primary objective is to address the inversion problem inherent to $n \gg m$ mappings by employing the differentiable properties of the MLM to guide gradient-based search strategies. Furthermore, we intend to validate the framework on a broader suite of real-world problems characterized by varying degrees of constraint complexity, extending beyond standard benchmarks to deployment in industrial engineering applications.

Ultimately, PGDS represents a fundamental paradigm shift in Many-Objective Optimization. By moving beyond opaque, black-box optimization and providing geometrically grounded, context-aware explanations, we pave the way for transparent and trustworthy decision-making in high-dimensional spaces.

\bibliographystyle{splncs04}
\bibliography{references}

\begin{thebibliography}{10}
\providecommand{\url}[1]{\texttt{#1}}
\providecommand{\urlprefix}{URL }
\providecommand{\doi}[1]{https://doi.org/#1}

\bibitem{abed2026evomapx}
Abed-Alguni, B.H.: Evomapx: An explainable framework for metaheuristic optimization algorithms. Expert Systems with Applications  \textbf{298},  129514 (2026). \doi{https://doi.org/10.1016/j.eswa.2025.129514}, \url{https://www.sciencedirect.com/science/article/pii/S095741742503129X}

\bibitem{bentley1975multidimensional}
Bentley, J.L.: Multidimensional binary search trees used for associative searching. Commun. ACM  \textbf{18}(9),  509–517 (Sep 1975). \doi{10.1145/361002.361007}

\bibitem{deb_nsgaII_2002}
{Deb}, K., {Pratap}, A., {Agarwal}, S., {Meyarivan}, T.: A fast and elitist multiobjective genetic algorithm: [nsga-ii]. IEEE Transactions on Evolutionary Computation  \textbf{6}(2),  182--197 (2002). \doi{10.1109/4235.996017}

\bibitem{deb2014nsga}
Deb, K., Jain, H.: An evolutionary many-objective optimization algorithm using reference-point-based nondominated sorting approach, part i: solving problems with box constraints. IEEE Transactions on Evolutionary Computation  \textbf{18}(4),  577--601 (2014)

\bibitem{deb2002nsga2}
Deb, K., Pratap, A., Agarwal, S., Meyarivan, T.: A fast and elitist multiobjective genetic algorithm: {NSGA-II}. IEEE Transactions on Evolutionary Computation  \textbf{6}(2),  182--197 (2002)

\bibitem{deb_sundar_2006}
Deb, K., Sundar, J.: Reference point based multi-objective optimization using evolutionary algorithms. In: Proceedings of the 8th Annual Conference on Genetic and Evolutionary Computation. p. 635–642. GECCO '06, Association for Computing Machinery, New York, NY, USA (2006). \doi{10.1145/1143997.1144112}

\bibitem{fieldsend2016visualising}
Fieldsend, J.E.: Visualising high-dimensional pareto relationships in two-dimensional scatterplots. In: International Conference on Evolutionary Multi-Criterion Optimization. pp. 258--272. Springer (2016)

\bibitem{Huband2011}
Huband, S., Hingston, P., Barone, L.: {A Review of Multi-objective Test Problems and a Scalable Test Problem Toolkit}. IEEE TEC  \textbf{10}(2006),  477--506 (2011)

\bibitem{lundberg2017unified}
Lundberg, S.M., Lee, S.I.: A unified approach to interpreting model predictions. In: Proceedings of the 31st International Conference on Neural Information Processing Systems. p. 4768–4777. NIPS'17, Curran Associates Inc., Red Hook, NY, USA (2017)

\bibitem{zhou2024evolutionary}
Mei, Y., Chen, Q., Lensen, A., Xue, B., Zhang, M.: Explainable artificial intelligence by genetic programming: A survey. IEEE Transactions on Evolutionary Computation  \textbf{27}(3),  621--641 (2023). \doi{10.1109/TEVC.2022.3225509}

\bibitem{meijer2025explainable}
Meijer, C., Bos, E.G.P.: Explainable embeddings with distance explainer. Journal of Machine Learning Research  \textbf{1},  1--33 (2025), arXiv:2505.15516v1 [cs.LG]

\bibitem{miettinen1999nonlinear}
Miettinen, K.: Nonlinear Multiobjective Optimization. Kluwer Academic Publishers, Boston, MA (1999)

\bibitem{misitano2024exploring}
Misitano, G.: Exploring the explainable aspects and performance of a learnable evolutionary multiobjective optimization method. ACM Transactions on Evolutionary Learning  \textbf{4}(1),  1--39 (2024)

\bibitem{misitano2022}
Misitano, G., Afsar, B., L\'{a}rraga, G., Miettinen, K.: Towards explainable interactive multiobjective optimization: R-ximo. Autonomous Agents and Multi-Agent Systems  \textbf{36}(2) (Oct 2022). \doi{10.1007/s10458-022-09577-3}

\bibitem{NAGAR2023101202}
Nagar, D., Ramu, P., Deb, K.: Visualization and analysis of pareto-optimal fronts using interpretable self-organizing map (isom). Swarm and Evolutionary Computation  \textbf{76},  101202 (2023). \doi{https://doi.org/10.1016/j.swevo.2022.101202}

\bibitem{petsiuk2018rise}
Petsiuk, V., Das, A., Saenko, K.: Rise: Randomized input sampling for explanation of black-box models. arXiv preprint arXiv:1806.07421  (2018)

\bibitem{ribeiro2016should}
Ribeiro, M.T., Singh, S., Guestrin, C.: "why should i trust you?": Explaining the predictions of any classifier. In: Proceedings of the 22nd ACM SIGKDD International Conference on Knowledge Discovery and Data Mining. p. 1135–1144. KDD '16, Association for Computing Machinery, New York, NY, USA (2016). \doi{10.1145/2939672.2939778}, \url{https://doi.org/10.1145/2939672.2939778}

\bibitem{deSouza2013mlm}
de~Souza~Junior, A.H., Corona, F., Miche, Y., Lendasse, A., Barreto, G.A., Simula, O.: Minimal learning machine: A new distance-based method for supervised learning. In: Proceedings of the International Work-Conference on Artificial Neural Networks (IWANN). Lecture Notes in Computer Science, vol.~7902, pp. 408--416. Springer (2013)

\bibitem{tusar2015visualization}
Tu{\v{s}}ar, T., Filipu{\v{c}}, B.: Visualization of pareto front approximations in evolutionary multiobjective optimization: A critical review and the prosection method. IEEE Transactions on Evolutionary Computation  \textbf{19}(2),  225--245 (2015)

\bibitem{zhang2007moead}
Zhang, Q., Li, H.: Moea/d: A multiobjective evolutionary algorithm based on decomposition. IEEE Transactions on Evolutionary Computation  \textbf{11}(6),  712--731 (2007)

\bibitem{zhen2017objective}
Zhen, L., Li, M., Cheng, R.: Objective reduction for visualising many-objective solution sets. Information Sciences  \textbf{418},  478--494 (2017)

\bibitem{zou2019visual}
Zou, J., Sun, Y., Li, M.: Visualizing the pareto front of many-objective optimization problems using t-sne. In: 2019 IEEE Congress on Evolutionary Computation (CEC). pp. 2456--2463. IEEE (2019)

\end{thebibliography}

\end{document}